\documentclass[letterpaper]{psa2023npic_hmit}
\usepackage{tabls}
\usepackage{cites}
\usepackage{epsf}
\usepackage{bm}
\usepackage{appendix}
\usepackage{ragged2e}
\usepackage[top=1in, bottom=1in, left=1in, right=1in]{geometry}
\usepackage{enumitem}
\setlist[itemize]{leftmargin=*}
\usepackage{caption}
\usepackage{amssymb}

\newcommand{\R}{\mathbb{R}}
\title{AI Enabled Neutron Flux Measurement and Virtual Calibration\\ in Boiling Water Reactors}
\author{%
  \textbf{~A. Tunga$^1$, ~J. Heim$^{1,2}$, ~M. Mueterthies$^1$, ~J. Thomas Gruenwald$^1$, ~J. Nistor$^{1,2}$} 
  \footnote{Corresponding Author | \texttt{jonathan@bwailabs.com} } \\
  \vspace{2pt} \\
  $^1$Blue Wave AI Labs, West Lafayette, Indiana \\ 
  $^2$Department of Physics and Astronomy, Purdue University, West Lafayette, Indiana \\ 
  \vspace{2pt}}
\newcommand{\authorHead}{Authors' names, use et al.~if more than three}
\newcommand{\shortTitle}{Short One Line Paper Title}
\begin{document}
\maketitle
\justify 
\parskip 6pt plus 1 pt minus 1 pt

\begin{abstract}
Accurately capturing the three-dimensional power distribution within a reactor core is vital for ensuring the safe and economical operation of the reactor, compliance with Technical Specifications, and fuel-cycle planning (safety, control, and performance evaluation). Offline (that is, during cycle planning and core design), a three-dimensional neutronics simulator is used to estimate the reactor’s power, moderator, void, and flow distributions, from which margin to thermal limits and fuel exposures can be approximated. Online, this is accomplished with a system of local power range monitors (LPRMs) designed to capture enough neutron flux information to infer the full nodal power distribution. Certain problems with this process, ranging from measurement and calibration to the power adaption process, pose challenges to operators and limit the ability to design reload cores economically (e.g., engineering in insufficient margin or more margin than required). Artificial intelligence (AI) and machine learning (ML) are being used to solve the problems to reduce maintenance costs, improve the accuracy of online local power measurements, and decrease the bias between offline and online power distributions, thereby leading to a greater ability to design safe and economical reload cores. We present ML models trained from two deep neural network (DNN) architectures, SurrogateNet and LPRMNet, that demonstrate a testing error of 1.1\% and 3.0\%, respectively. Applications of these models can include virtual sensing capability for bypassed or malfunctioning LPRMs, on-demand virtual calibration of detectors between successive calibrations, highly accurate nuclear end-of-life determinations for LPRMs, and reduced bias between measured and predicted power distributions within the core. 

\vspace{6pt}
\end{abstract}
\vspace{6pt}

\section{INTRODUCTION} 
In boiling water reactors (BWRs), an array of fixed in-core detectors (LPRMs) \cite{morgan1970core} is used and supplemented with movable in-core detectors (TIPs), to measure the local power distribution. These detectors provide perhaps the most fundamental set of measurements within a reactor core. Accurately capturing the full power distribution within a reactor core is vital for ensuring the safe and economical operation of the reactor, compliance with Technical Specifications, and fuel-cycle planning. Offline (that is, during cycle planning and core design), a three-dimensional neutronics simulator is used to estimate the reactor’s power, moderator, void, and flow distributions, from which margin to thermal limits and fuel exposures can be approximated. Online, this is accomplished with the system of LPRMs to capture enough neutron flux information to infer the full nodal power distribution.

Certain problems with this process pose challenges to operators and limit the ability to design reload cores economically (e.g., engineering in insufficient margin or more margin than required). These include (i) lack of visibility into local power distribution when an LPRM is bypassed or otherwise inoperable, (ii) incorrect and infrequent detector calibrations, (iii) premature or overdue replacement of LPRMs that have reached their end-of-life (EOL), and (iv) inaccurate power adaption that has downstream effects on perceived margin to operating limits. There is a clear opportunity to apply artificial intelligence (AI) and machine learning (ML) to improve nuclear power plant (NPP) efficiency and reduce costs \cite{nuclearnews}. AI can be used over a wide range of nuclear plant operations, from predicting component lifetimes and evaluating asset health to understanding core dynamics for more accurate reload planning and economical fuel purchasing. In this work, we develop multiple deep neural network (DNN) architectures to improve the accuracy of online local power measurements, and decrease the bias between offline and online power distributions, thereby leading to a greater ability to design safe and economical reload cores.  

There is a vast amount of historical data available, as NPPs routinely collect surveillance and diagnostic data throughout the various plant systems.  For the LPRM system, this includes real-time collection of raw and procesed signals from the LPRM detectors (with a collection rate on the order of 1Hz). Complementary to these measurement data are the core statepoints as determined by the nodal core simulator and the operating conditions of the core, with similar collection frequency. In this paper, we make use of this data to develop two classes of models. SurrogateNet utilizes readings from other LPRMs to predict any other LPRM's reading, and LPRMNet uses only known (and forecastable) reactor state variables, such as rod patterns, offline nodal power, core flow, and others to predict the LPRM values.

\section{BACKGROUND}
\subsection{LPRM}
\begin{figure}[!htb]
  \centering
  \includegraphics[scale=0.35]{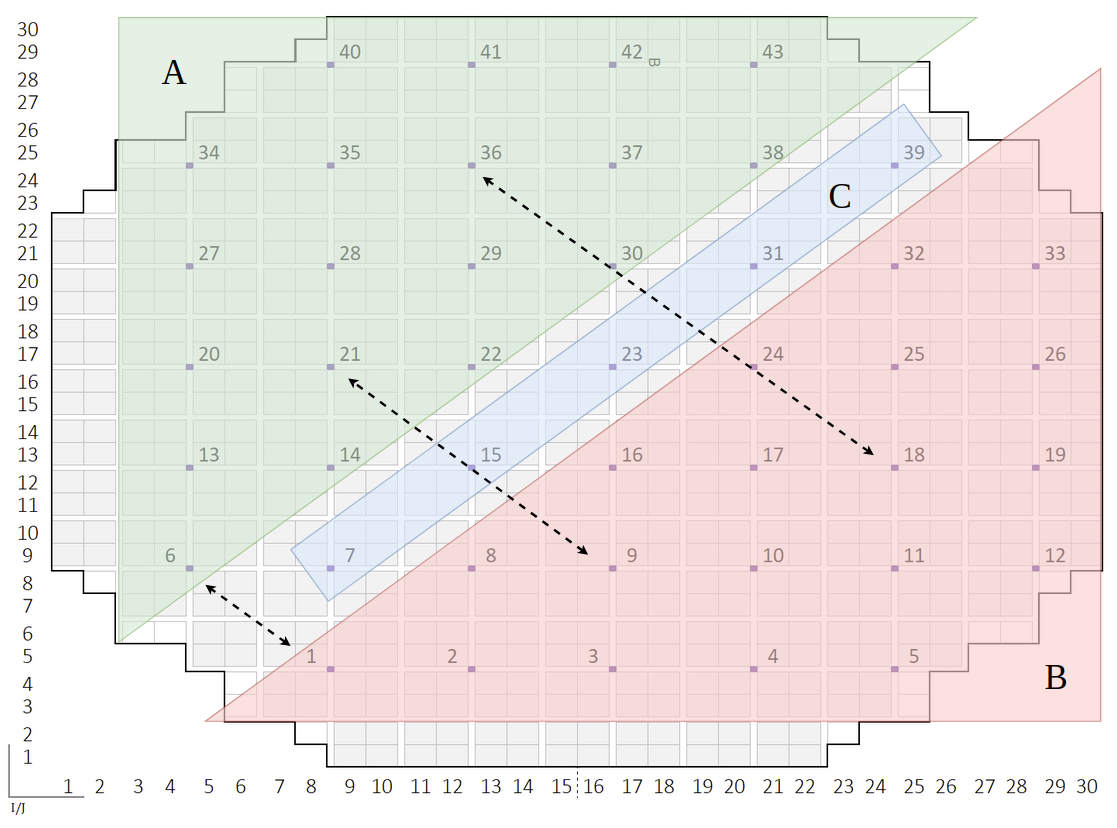}
  \caption{Distribution of LPRMs. All the LPRMs in sets $\mathcal{A}$ and $\mathcal{B}$ have symmetric partners, a few of them are indicated by the dashed arrows. The LPRMs in the set $\mathcal{C}$ do not have symmetrical partners. }
  \label{fig:react}
\end{figure}

The Local Power Range Monitoring (LPRM) System \cite{morgan1970core} provides signals proportional to the local neutron flux, and the individual LPRMs measure the local flux at various radial and axial locations within the core. There are typically 43 radially located LPRM assemblies (strings) in a large BWR, with each assembly containing four detectors spaced at 36-inch intervals. Each LPRM assembly is adjacent to a hollow dry instrument tube for the Traversing Incore Probe (TIP) System. This system is used to periodically calibrate the LPRMs to correct for instrument drift and to collect more spatially-resolved axial power distributions (referred to as traces) throughout the core. Mostly, the distribution of LPRMs is such that they are symmetric across a diagonal axis. In Figure \ref{fig:react}, the LPRMs in sets $\mathcal{A}$ and $\mathcal{B}$ are symmetric, and set $\mathcal{C}$ forms the symmetry axis. The readings of the symmetric partners are very similar, given symmetric operation of the fuel cycle. As a representative example, Figure \ref{fig:data} shows the readings of 1A and 6A, which are symmetric partners.

To be able to reliably construct the axial flux distribution of the core, the four detectors are spaced at three-foot intervals. The lowest detector, Detector A, is located 18 inches above the bottom of the active fuel. The remaining detectors are spaced 36 inches apart with the D detector located 24 inches from the top of the fuel assembly. 

The LPRM detector is a miniature fission chamber, with the case and collector fabricated from titanium and insulated from one another by a ceramic material. The inner surface of the case is coated with a U$_3$O$_8$ coating which contains several isotopes of uranium. This will include 18\% U-235, 78\% U-234 and 4\% various isotopes of uranium (primarily U-238). The fill gas used in the LPRM detector is argon with an internal pressure of 14.7 psia (atmospheric). Thermal neutrons impinging on the detector have a high probability of causing uranium atoms to fission. The resulting fission process releases fission fragments, neutrons, and gamma radiation into the detector volume. This causes ionization of the gas and an electrical discharge between cathode and anode. Gamma radiation from sources external to the detector can also cause ionization within the detector.

The mixed uranium isotopes (U-235 and U-234) in the U$_3$O$_8$ coating provides two functions. The U-235 has a high probability of fission which causes the primary ionization of the argon gas within the LPRM detector. If U-235 were the only uranium isotope in the LPRM detector, the uranium coating would be rapidly depleted. This would result in sensitivity of the detector decreasing rapidly at normal operating power levels. Earlier BWR models were equipped with this type of LPRM detector (non-regenerative). The addition of U-234 provides a method of replacing or regenerating U-235 lost by fission. U-234 has a very low probability of fission but a high probability of adsorption and transmutation to U-235. 

Since the LPRM detectors permanently reside in the core during full-power operation, they receive an accumulating high neutron flux exposure throughout their lifetime. The quantity of uranium atoms available for reaction depletes as LPRM detector exposure increases. This causes a decrease in the sensitivity of the detector. To compensate for this decreased sensitivity, each LPRM channel is periodically calibrated using the TIP System. The exposure of each LPRM is tracked and forecasted to ensure replacement scheduling before it reaches EOL (the LPRMs can only be replaced during refueling outages). The neutron-rich environment can also lead to mechanical failures and signal issues, requiring the malfunctioning unit to be bypassed by the operator, and the detector signal would no longer be available to downstream calculations. 

In this work, we have used deep learning networks to develop two highly accurate models, SurrogateNet and LPRMNet, that can predict the value of any LPRM at any time. These models provide a means for virtual calibration, virtual measurement for off-line LPRMs and for predicting behavior in future core states, thereby enabling more accurate EOL determinations. 

\subsection{DEEP LEARNING}
Recent advances in AI like image recognition, autonomous driving, voice recognition, and language models are possible due to Deep Learning (DL). In DL, a large model learns to perform a task (usually classification or regression) directly from the data. DL models can learn intricate structures in high-dimensional data, and representations of data required to perform a given task \cite{lecun2015deep}. In this paper, we use a form of DL, called supervised learning, with a large dataset consisting of input features and targets. In the DL model's training process, it has access to the input features and desired targets. The DL model starts off with random adjustable weights, which are applied to the input features, to produce an output. The model's output is compared with the desired output, and an error measure (loss) is computed between the two. In the training process, the weights of the model are adjusted to minimize the error between the predicted output and the desired targets. Usually, the number of these adjustable weights, also called neurons, is on the order of millions. There are many DL model architectures that vary based on the representation of neurons and are used for different tasks. In this paper, we use two such architectures – Fully Connected Networks, and Convolutional Neural Networks (CNNs), the details of which are presented in the following sections. 

\subsubsection{Fully connected networks}
\begin{figure}[!htb]
  \centering
  \includegraphics[scale=0.65]{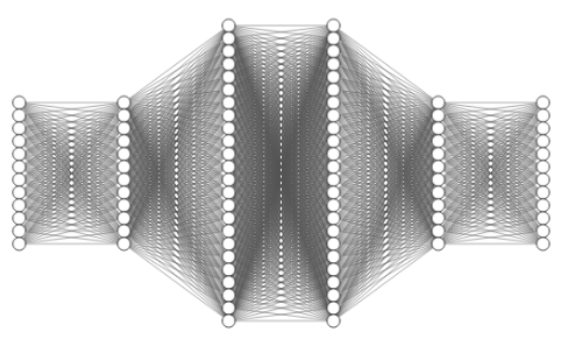}
  \caption{SurrogateNet and arrangement of LPRM assembly. The number of neurons in the six fully connected layers has been scaled down for illustration. }
  \label{fig:surr}
\end{figure}

A fully-connected neural network is the simplest DL model, where it takes a fixed-size one-dimensional vector as an input, performs non-linear transformation, and outputs a fixed-size one-dimensional network. A fully connected network has an input layer, an output layer, and multiple hidden layers. It is called fully connected because every neuron in one layer is connected to every other neuron in the subsequent layer. Figure \ref{fig:surr}, shows the architecture of one of the fully connected networks used in this paper.  

\subsubsection{Convolutional neural networks}
Recently, CNNs have shown tremendous performance improvements in image processing tasks such as classification \cite{krizhevsky2017imagenet}, segmentation \cite{he2017mask}, and detection \cite{redmon2016you}. CNNs are particularly good at processing data with multiple array-like structures. For example, images made of three 2D arrays, one for each channel -- red, green, and blue. The main idea behind CNNs is local connectivity -- each neuron is only connected to a local region instead of all neurons \cite{lecun2015deep}. In a typical CNN, a kernel with learnable weights slides over the array, and computes the dot product between its weights and the values in an array, in a particular region. The dot product produces a value indicative of how strongly a specific pattern or feature is present in that region. In most cases, multiple features might be required to learn a specific task, so multiple kernels are learned, and dot products are computed across the image to produce feature maps. These feature maps are stacked and passed through another set of convolutional layers, depending on the depth of the network. Usually, the initial layers focus on the local relationships, while the later layers focus on the global relationships in data. Since the structure of data from a NPP core is very similar to images, with multiple 2D arrays stacked on top of each other (images usually have 3 arrays stacked, while the data from NPPs usually has 25 nodal layers stacked), CNNs are well suited to process this data. Also, recent work \cite{shriver2022scaling} has shown success in using CNNs to predict pin powers in pressurized water reactors. 

\subsubsection{Attention}
One of the important properties of CNNs is local connections, which help in reducing the number of parameters in the network. However, they introduce a major challenge: modeling long-range relationships. Recent works have proposed methods like using multi-scale dilatation convolutions to aggregate contextual information \cite{chen2017deeplab} and pyramid pooling \cite{zhao2017pyramid} to capture contextual information. However, the dilation-based methods focus exclusively on local regions and cannot produce rich contextual information. Similarly, pyramid pooling-based methods do not account for the fact that different pixels have different contextual dependencies. To introduce rich and pixel-wise contextual information, recent works \cite{vaswani2017attention,Huang_2019_ICCV} utilize the self-attention mechanism. Using the attention mechanism enriches the CNN with non-local and long-range contextual information. However, the attention mechanism, in its native form is computationally expensive. Based on \cite{wang2020axial}, we use axial-attention to enable efficient computation and to model long-range contextual information. 

\section{DATA} 
The data used for all experiments discussed in this work are from a currently operating large BWR-4 (with Mark I containment) NPP. The dataset includes measured LPRM readings, measured core parameters, and parameters computed by a commercial core simulator. The parameters used as input features are Nodal Power ($NP \in \R ^{H \times W \times D}$), Nodal Blade Depletion ($NBD \in \R ^{H \times W \times D^{'}}$), and Rod Pattern ($RP \in \R ^{H \times W}$). We also use one-dimensional parameters -- thermal power, core inlet subcooling, and core flow. The measured LPRM values are the targets that the models predict using the input features.

\begin{figure}[!htb]
  \centering
  \includegraphics[scale=0.37]{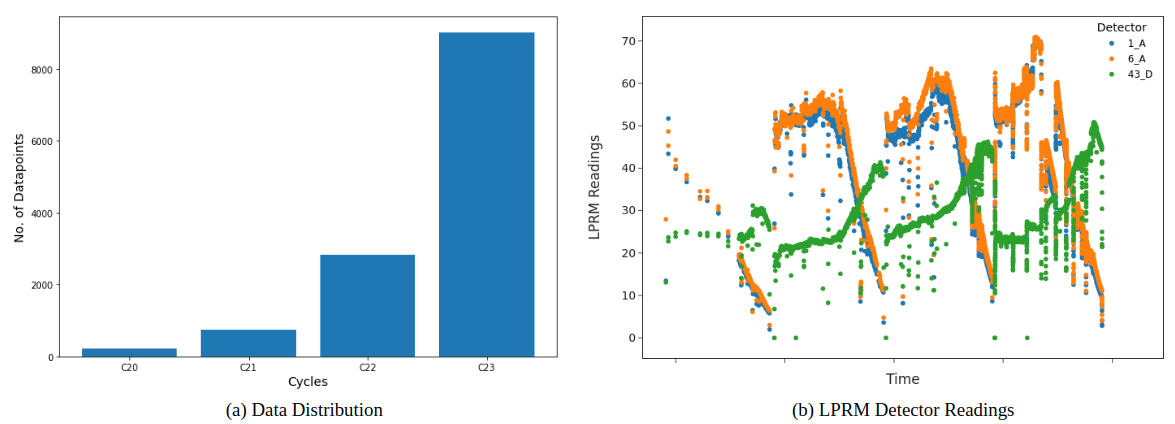}
  \caption{Overview of the data. }
  \label{fig:data}
\end{figure}

The frequency and the number of available LPRM measurements vary with each cycle, with the latest cycles having more frequent data collection. The distribution of the data across different cycles is shown in Figure \ref{fig:data}(a). Measured LPRM values for a few of the detectors, 1A, 6A, and 43D, are depicted in Figure \ref{fig:data}(b). The dataset is cleaned by removing invalid outliers, and transient data by excluding all points where the thermal power is less than 90\% of the rated power. Some features are preprocessed to impart more information to the models. For example, the rod pattern, alone, indicates the amount of insertion of the control rods yet does not take into account depletion of the control blades. Highly depleted regions of blades will be less effective than non-depleted regions. To account for this depletion, we combine nodal blade depletion and rod pattern to derive a new feature Rod Variable (RV) as follows: 
\begin{itemize}\itemsep6pt \parskip0pt \parsep0pt
\item Nodalize rod pattern to intermediate rod variable ($RV^{'}$) by representing the insertion by 1s and 0s in the $D^{'}$ dimension. For example, if at a given location, the rod is 50\% inserted, then half of the values in $D^{'}$ dimension at that location will have 1s, rest 0s. 

\item Weight the nodalized array with segment depletion: 
\begin{equation}
  \label{nbd}
  s_{i}= 1 - NBD_{i}, \;
  RV_{i} = s_i * RV^{'}_i,
\end{equation}
where $NBD_{i} \in NBD$, $RV^{'}_i \in RV^{'}$ and $RV_i \in RV$ at position $i$, and $s_i$ represents the fraction of the blade not depleted at position $i$. In a typical large-size BWR reactor $H=W=30$, $D=25$, and $D^{'}=24$.
\end{itemize}

\section{PROPOSED ARCHITECTURE} 
\label{sec:first}
In this section, we discuss the architectures and modeling methodology of the SurrogateNet and LPRMNet models, which are developed for real-time predictions and more accurate offline predictions, respectively.

\subsection{SurrogateNet} 
SurrogateNet is a neural network-based architecture that uses information from multiple LPRM detectors resident in the core to predict the value of any other LPRM detector in the core contemporaneously. We exploit the symmetry in the core to predict the LRPM detector values. We divide the LPRMs in the core into three distinct sets---$\mathcal{A}$ (green) with $19$ strings and $76$ detectors, $\mathcal{B}$ (red) with $19$ strings and $76$ detectors and $\mathcal{C}$ (blue) with $5$ strings and $20$ detectors, as shown in Figure \ref{fig:react}. Every detector in set $\mathcal{A}$ has a symmetric partner in set $\mathcal{B}$, as indicated by the dotted lines in Figure \ref{fig:react}. The set $\mathcal{C}$ forms the line of symmetry, and thus the detectors in $\mathcal{C}$ do not have symmetrical partners. 

To predict the values of the detectors in sets $\mathcal{A}$ and $\mathcal{B}$, one of the sets is used as an input set and the other is used as the output set, respectively. The input set of $76$ detectors is passed through six fully connected layers. For the fully connected layers, GELU \cite{hendrycks2016gaussian} is used as the non-linear activation function, and for each layer batch normalization is performed. The optimal parameters for the network are identified through hyper-parameter optimization. The same architecture is used for the model where the input and output sets are reversed. Using the two models, $152$ of the $172$ detectors can be calculated in real-time. To account for the situations where the detectors are bypassed or otherwise deemed faulty, data augmentation is performed on the fly when training the network, which is explained in Section \ref{sec:implement}. 

The detectors in set $\mathcal{C}$ do not have symmetrical partners, so a modified model and different inputs have been used to train the model. For every detector in $\mathcal{C}$, all but the target detector is used as inputs to the model ($N$-1 detectors are used as inputs, where $N$ is the total number of detectors in the core). The $N$-1 detectors are passed through six fully connected layers before predicting the final detector value. The activation function and batch normalization are similar to the previously defined models. As these models are based on simple fully connected networks, they can be used for real-time predictions, even on a CPU, thereby enabling real-time monitoring and diagnostics.  

\subsection{LPRMNet} 
\label{sec:second}
LPRMNet is a CNN-based model, which uses the reactor state data to predict the LPRM values. In the implementation of LPRMNet, we extract information from $NP$ and $RV$ separately using CNNs and use fully connected networks to extract the information from the scalar variables. $NP$ and $RV$ are separately passed to individual CNNs, and the output from the individual CNNs is stacked to form a feature map {$L$}. The scalar features are passed through two fully connected layers to produce a feature vector {$S$}.

\begin{figure}[!htb]
  \centering
  \includegraphics[scale=0.4]{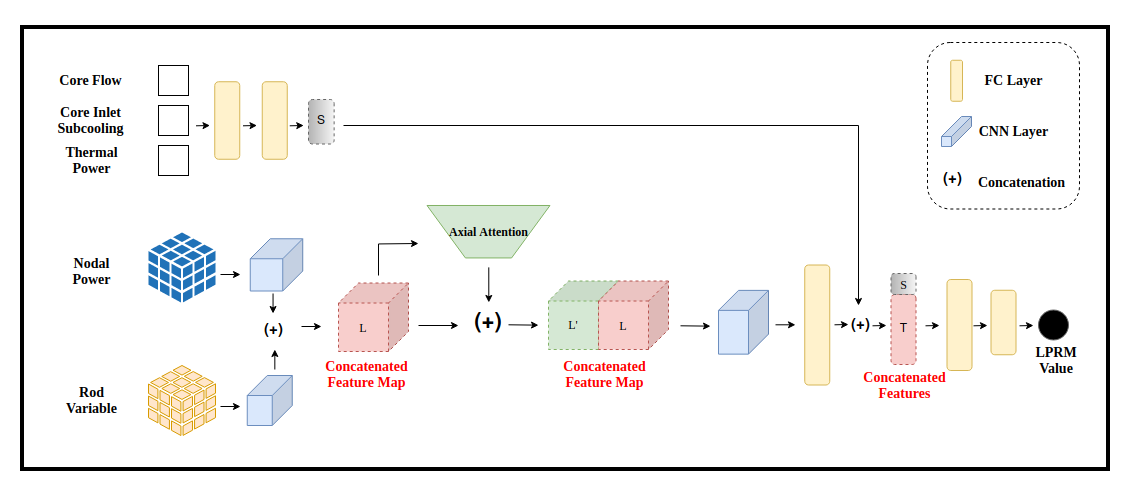}
  \caption{Overview of the proposed LPRMNet. }
  \label{fig:arch}
\end{figure}\vspace{-5pt}

The feature map $L$ is passed through the axial-attention \cite{wang2020axial} module to impart contextual knowledge across the horizontal and vertical axes. Axial-attention enables a global receptive field, compared to the receptive field achieved by only using CNNs. Similar to \cite{Huang_2019_ICCV}, in the axial-attention module, $1 \times 1$ convolution is applied on feature map $L$ to generate feature maps $Q$ and $K$ of the same spatial dimensions as $L$. From $Q$ and $K$, an attention map $M$ is obtained via the Affinity operation. From every spatial position of $Q$, a vector $Q_j$ is extracted along the channel dimension of $Q$. From $K$, a feature set  $\phi_j$ is extracted from the same row or column as $j$. The Affinity operation is defined as:
\begin{equation}
  \label{first_equation}
  a_{i,j}= Q_j\phi_{i,j}^T,
\end{equation}
where $a_{i,j} \in A$, and a softmax layer is applied to $A$ to get the attention map $M$. Further, a $1 \times 1$ convolution is again applied on feature map $L$, to generate $V$. Similar to $Q_j$, from every spatial location of $V$, a vector $V_j$ is obtained, and a set of features which are in the same row or column of $j$ is obtained as $\Omega_j$. Using the Aggregation operation, the contextual information is collected:
\begin{equation}
  \label{second_equation}
  L_j^{'} = \Sigma_i M_{i,j} \Omega_{i,j} + L_{j} ,
\end{equation}
where $L_j^{'}$ is a feature vector in $L^{'}$ at position $j$. $L^{'}$ is the contextual information-rich feature map derived from $L$. The output from the axial-attention module, $L^{'}$, is concatenated with the feature map $L$. Further, the concatenated feature map is flattened into a single vector and passed through a fully connected network to obtain a dense vector $T$. The vectors $T$ and $S$ are concatenated and passed through a regression stack of fully connected layers to finally predict the value of the LPRM detector. Figure \ref{fig:arch} shows the architecture of LPRMNet. 

\subsection{IMPLEMENTATION DETAILS} 
\label{sec:implement}
The proposed models - SurrogateNet and LPRMNet - have been implemented in PyTorch \cite{paszke2019pytorch}, and trained using mean squared error loss. For training SurrogateNet, AdamW \cite{loshchilov2017decoupled} optimizer is used with a weight decay of $0.01$. 1-cycle policy \cite{smith2019super} is used for the learning rate with $max\_lr = 0.005 $. Since the LPRM  detectors are predicted using other LPRM detectors, it is possible that the model might not be robust when some of the detectors are bypassed when faulty (they read 0). To overcome this challenge, while training the SurrogateNet, with a probability of 20\%, the input detectors' are replaced with 0s. We observed that by randomly replacing detector values with 0s, the model is stable during inference even when one or more detectors are bypassed.  

LPRMNet is trained with AdamW \cite{loshchilov2017decoupled} optimizer with a weight decay of $0.01$. Similarly, a 1-cycle policy is used for the learning rate with $max\_lr = 0.08 $. Since each detector has an individual LPRMNet model, a total of 172 different models have been trained. To speed up the training, 4 models have been trained in parallel on a single GPU. All the models have been trained on a single 24 GB NVIDIA A30 GPU.

\section{EXPERIMENTS AND RESULTS}
\subsection{DATA SPLITS}
\subsubsection{SurrogateNet data split}
SurrogateNet is developed to predict real-time LPRM readings, so it is beneficial to evaluate it across all the cycles. From the entire dataset, 70\% of the data is used for training, and 20\% of the data is used for validation. The remaining 10\% of the data is used for independent testing and is never seen in the training or hyper-parameter optimization process. All the results in  Table \ref{table:surr_res} are reported on the independent test set.  

\subsubsection{LPRMNet data split}
LPRMNet predicts the LPRM readings from offline core parameters. The model predicts the LPRM values for new fuel cycles, which will aid in better planning of the fuel cycles. To test the robustness of such a model, it is crucial to test the model's performance on a new cycle, which has not been used in training.  Therefore, for training LPRMNet, we hold out cycle 22 (Fig. \ref{fig:data}) from the training set and use the rest of the data for training. From cycle 22, we use 50\% of the data in validation, and the rest is used as an independent test set. Testing a model on a new cycle shows the model's ability in generalizing not only to new data but also to new data distributions. Ideally, a complete cycle in the future should be held out for testing, but since cycle 23 contains more than 80\% of the available data (Figure \ref{fig:data}), cycle 22 is used for testing - it is not feasible to train the model without cycle 23 data.   

\subsection{RESULTS}

\vspace{-15pt}\begin{table}[!htb]
  \centering
  \caption{\bf  SurrogateNet results on the test set.}
  \label{table:surr_res} 
  \begin{tabular}{|c|c|c|c|c|c|} \hline 
  & \multicolumn{2}{c|}{\textbf{SurrogateNet}} & \multicolumn{3}{c|}{\textbf{LPRMNet}} \\\hline
    & \textbf{Mean RMSE} & \textbf{Max RMSE} & \textbf{Mean RMSE} & \textbf{Max RMSE} & \textbf{Core Simulator} \\\hline
    \ Overall &  $5.0~10^{-1}$& $8.8~10^{-1}$ &  $1.35$ & $2.28$ & $3.68$       \\ \hline
    \ A Level Detectors& $5.8~10^{-1}$& $8.8~10^{-1}$& $1.41$& $2.13$ & $3.45$ \\ \hline 
    \ B Level Detectors& $4.8~10^{-1}$& $5.9~10^{-1}$ & $1.18$& $1.86$ & $3.84$\\ \hline 
    \ C Level Detectors& $4.6~10^{-1}$& $5.9~10^{-1}$ & $1.32$& $2.10$ & $3.95$ \\ \hline 
    \ D Level Detectors& $4.8~10^{-1}$& $6.2~10^{-1}$ & $1.5$& $2.28$ & $3.56$\\ \hline 
  \end{tabular}
\end{table}\vspace{-5pt}

\begin{figure}[!htb]
  \centering
  \includegraphics[scale=0.4]{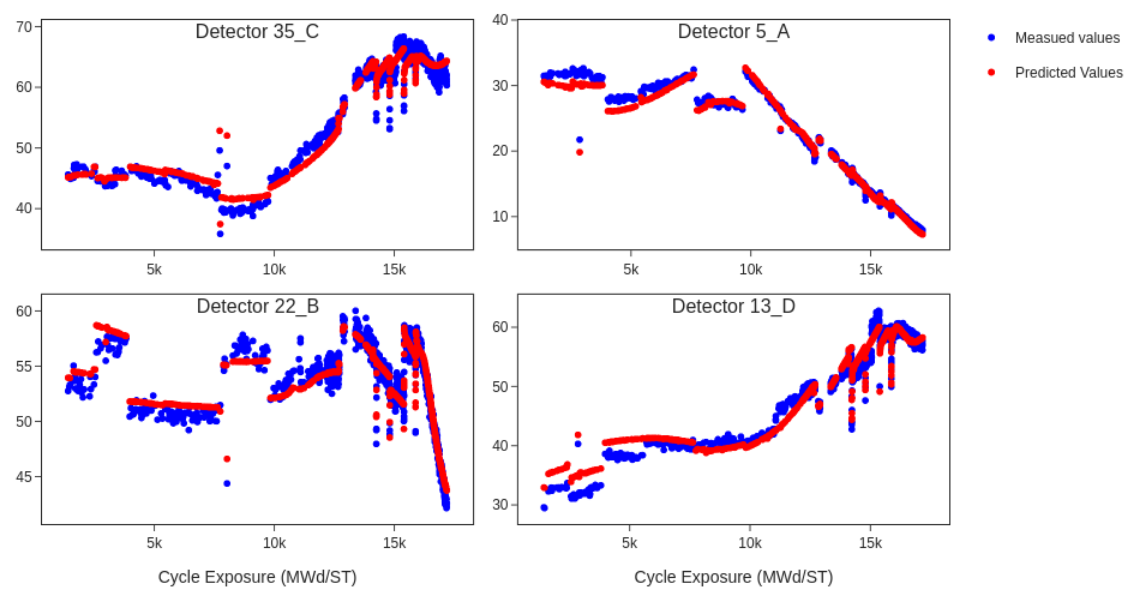}
  \caption{Visualization of few results from LPRMNet. The values in blue are the measured LPRM values and the values in red are the model predictions. The $y$-axis represents the LPRM values. }
  \label{fig:fin_res}
\end{figure}

To evaluate the models, the Root Mean Square Error (RMSE) between the model predictions and the detector measurements is calculated for the test sets, for all the detectors. The average RMSE, as well as the maximum MSE for all the detectors, can be seen in Table \ref{table:surr_res}, for SurrogateNet and LPRMNet, respectively. In addition to the average RMSE, the average RMSE at each detector level - A, B, C, and D is also shown. We compare our results with the results from a commercial core simulator and show a consistent reduction in errors on each of the detector levels for the test set (Table \ref{table:surr_res}). In Figure \ref{fig:fin_res}, we visually show how the model's predictions compare with the actual measurements.

On comparing the results from the SurrogateNet and the LPRMNet, though it might appear that the results from SurrogateNet are better than the results from LPRMNet, it should be noted that the test sets are vastly different between the two. In the case of SurrogateNet, although the test data are unseen, they come from the same distribution as the training set. However, in the case of LPRMNet, the test data is both unseen and from a new distribution as compared to the training data. Also, SurrogateNet has access to online core conditions, when the measurements are taken. The different testing methodologies are intentionally chosen based on the use cases of both models. Though both the models predict the LPRM detector values, they are useful in different situations - the SurrogateNet for online predictions and virtual readings, while the LPRMNet for offline and future predictions.

\section{CONCLUSIONS}
In this paper we demonstrate that by using the available data from a NPP, spanning 4 fuel cycles, we can reliably predict the values of the LPRM detectors using deep learning modeling methodologies. Two different models were developed: (i) an online model, which uses the values of other LPRM detectors to predict the value of a given LPRM detector with a mean RMSE of $0.5$, and (ii) an offline prognostic model, which can be used for future planning, that uses core projection data to predict the values of the LPRM with a mean RMSE of $1.35$. This level of performance corresponds to an average percent error of 1.1\% and 3.0\% for type (i) and (ii) models, respectively. The models have shown more than 2$\times$ reduction in uncertainty when compared to the relevant commercial core simulator.

Applications of these models include virtual sensing capability for bypassed or malfunctioning LPRMs, on-demand virtual calibration of detectors between successive TIP calibrations, highly accurate nuclear end-of-life determinations for LPRM detectors, and reduced bias between measured and predicted power distributions within the core. The latter of these has direct relevance to reducing biases that exists between offline and online thermal limit determinations. Accurate predictions of core-wide and local power distributions are crucial to assuring that targeted margins to the operating limits are maintained. Deviation between measured performance and design predictions lead to operational challenges, such as unplanned power derates, premature coast down, or increased fuel costs due to loading more fuel than required for targeted energy production. The inability to accurately predict online thermal limits from offline methods has challenged core design and cycle management within the industry. The biases between online and offline methods originate, in part, from the absence of feedback available from the in-core instrumentation (i.e. LPRMs) during core design and other offline scenarios. The models of type (ii) presented in this work allow for a proxy to this online feedback during reload core design, thereby reducing the uncertainties commonly observed through current industry practices. 

Finally, it is worth noting that DL models perform very well when the testing and training data share the same distribution but may significantly underperform when they are from different distributions. Bridging this gap is an active area of research. In the setting described here, introducing new operating conditions such as changes in fuel design, or alternate reactivity control strategies, have the potential to change the data distribution of a fuel cycle substantially. Work is currently underway to develop methods, models, and training schemes to make the models robust to distribution shifts and achieve distribution generality.

\section*{ACKNOWLEDGEMENTS}

The authors would like to acknowledge the U.S. Department of Energy for supporting many industry-led efforts through various funding pathways established within the {\em U.S. Industry Opportunities for Advanced Nuclear Technology Development} (DE-FOA-0001817) program. This material is based upon work supported by the Department of Energy under award DE-NE0008930. This report was prepared as an account of work sponsored by an agency of the United States Government. Neither the United States Government nor any agency thereof, nor any of their employees, makes any warranty, express or implied, or assumes any legal liability or responsibility for the accuracy, completeness, or usefulness of any information, apparatus, product, or process disclosed, or represents that its use would not infringe privately owned rights. Reference herein to any specific commercial product, process, or service by trade name, trademark, manufacturer, or otherwise does not necessarily constitute or imply its endorsement, recommendation, or favoring by the United States Government or any agency thereof. The views and opinions of authors expressed herein do not necessarily state or reflect those of the United States Government or any agency thereof.


\setlength{\baselineskip}{12pt}
\bibliographystyle{psa2023npic_hmit}
\bibliography{psa2023npic_hmit}



\end{document}